\documentclass{article}

\newcommand{\mdp}{\mathcal{M}}
\newcommand{\sspace}{\mathcal{S}}
\newcommand{\aspace}{\mathcal{A}}
\newcommand{\saspace}{\mathcal{Z}}

\newcommand{\rawrew}{\mathcal{R}}

\newcommand{\rawtrns}{P}
\newcommand{\rew}{\rawrew}

\newcommand{\trns}{\rawtrns}

\newcommand{\dist}{{Dist}}

\usepackage{anomy}

\usepackage{times}
\usepackage{soul}
\usepackage{url}
\usepackage[hidelinks]{hyperref}
\usepackage[utf8]{inputenc}
\usepackage[small]{caption}
\usepackage{graphicx}
\usepackage{amsmath}
\usepackage{amsthm}
\usepackage{booktabs}
\usepackage{algorithmic}
\usepackage[switch]{lineno}

\usepackage{subcaption}

\usepackage{algorithm}

\usepackage{xcolor}

\usepackage{amsmath, amssymb}

\usepackage{amssymb}

\title{Off-Policy Actor-Critic with Sigmoid-Bounded Entropy for Real-World Robot Learning}

\author{
Xiefeng Wu$^1$
\and
Mingyu Hu$^1$
\and
Shu Zhang$^1$\\
\affiliations
$^1$Wuhan University\\
\emails
wuxiefeng@whu.edu.cn,
mingyuhu@whu.edu.cn,
00033521@whu.edu.cn
}

\begin{document}

\maketitle

\begin{abstract}
Deploying reinforcement learning in the real world remains challenging due to sample inefficiency, sparse rewards, and noisy visual observations. Prior work leverages demonstrations and human feedback to improve learning efficiency and robustness. However, offline-to-online methods need large datasets and can be unstable, while VLA-assisted RL relies on large-scale pretraining and fine-tuning. As a result, a low-cost real-world RL method with minimal data requirements has yet to emerge. We introduce \textbf{SigEnt-SAC}, an off-policy actor-critic method that learns from scratch using a single expert trajectory. Our key design is a sigmoid-bounded entropy term that prevents negative-entropy-driven optimization toward out-of-distribution actions and reduces Q-function oscillations. We benchmark SigEnt-SAC on D4RL tasks against representative baselines. Experiments show that SigEnt-SAC substantially alleviates Q-function oscillations and reaches a 100\% success rate faster than prior methods. Finally, we validate SigEnt-SAC on four real-world robotic tasks across multiple embodiments, where agents learn from raw images and sparse rewards; results demonstrate that SigEnt-SAC can learn successful policies with only a small number of real-world interactions, suggesting a low-cost and practical pathway for real-world RL deployment.
\end{abstract}

\section{Introduction}
Deep Reinforcement Learning (RL) holds the promise of enabling generalist robotic agents to perform different tasks in unstructured environments. However, the realization of this potential is severely hindered in real-world settings by expensive data collection~\cite{luo2024hilserl,rl-100}, noisy observations, and an unstable RL learning process~\cite{calql,cql}. To address these challenges, existing methods typically rely on pre-collected datasets to reduce exploration cost and stabilize training~\cite{rl-100,awac}, as well as human-in-the-loop intervention~\cite{luo2024hilserl} or RL-based fine-tuning of pretrained vision-language-action (VLA) policies~\cite{vla-3,vla-4,vla-5,vla-6} to improve task success rates.

We identify two primary barriers that prevent current methods from achieving efficient, learning-from-scratch execution: (1) Dependency on Heavy Priors: Offline-to-online approaches typically require massive, high-quality datasets to overcome distribution shifts. Acquiring such datasets for every new task is often prohibitively expensive and labor-intensive. (2) Limited Validation Under Realistic Dynamics and Noise: Existing evaluations are predominantly conducted on relatively stable robotic arm manipulation with multi-view camera setups that increase observation dimensionality and simplify state estimation. As a result, these methods are rarely stress-tested under noisy sensing or high-speed, highly dynamic motions, leaving their RL performance and robustness in such challenging regimes largely unverified. Thus, a practical RL framework with low cost in both data collection and real-world deployment remains elusive.

To address these challenges, we introduce \textbf{SigEnt-SAC}, a sample-efficient RL framework capable of learning from scratch without large-scale offline pre-training. It enables rapid convergence in sparse-reward tasks using only a single expert trajectory. SigEnt-SAC mitigates Q-value oscillations in conservative Q-learning and accelerates exploration through two key designs: (1) \textbf{Sigmoid-Bounded Entropy}, a novel entropy formulation for the tanh-squashed Gaussian policy that maps per-dimension surprisal through a sigmoid to produce a bounded, numerically stable entropy signal. This design prevents negative-entropy–dominated Q-learning updates that can amplify Q-value oscillations and drive the policy to optimize toward out-of-distribution (OOD) actions;
 and (2) \textbf{Gated Behavior Cloning (GBC)}, which penalizes only those policy actions that deviate substantially from dataset (expert) samples, thereby stabilizing policy optimization under Q-value oscillations and preventing unnecessary exploration.

We extensively evaluate \textbf{SigEnt-SAC} across two distinct settings: D4RL Adroit and Kitchen benchmarks~\cite{d4rl}, and diverse real-world robotic tasks. Our main contributions are summarized as follows:

\begin{itemize}
    \item \textbf{One-Shot Efficiency:} In the one-shot setting, \textbf{SigEnt-SAC} reaches \textbf{100\%} evaluation task success faster than state-of-the-art offline-to-online baselines; compared to conservative methods, it substantially reduces the data acquisition burden for sparse-reward learning and mitigates OOD action exploration.

    \item \textbf{Single-Camera Control in Dynamic Environments:} \textbf{SigEnt-SAC} supports control from a \emph{single} egocentric camera stream and still achieves stable policy convergence in \emph{moving} and visually dynamic environments, demonstrating robustness to partial observations and blurred visual inputs.

    \item \textbf{Cross-Embodiment Generalization:} \textbf{SigEnt-SAC} generalizes across diverse physical embodiments, including humanoid robots, quadrupedal robots, and autonomous mobile robots, demonstrating broad applicability.
\end{itemize}

\section{Related work}

\subsection{Offline-to-Online Reinforcement Learning}
Offline-to-Online RL aims to warmstart the learning process by pre-training policies on static datasets prior to online fine-tuning. Prominent methods such as IQL~\cite{iql}, CQL~\cite{cql}, AWAC~\cite{nair2020awac}, TD3+BC~\cite{td3bc}, and Cal-QL~\cite{calql} facilitate this transition by imposing policy constraints or value regularization to ensure the policy remains within the support of the offline data. Additionally, RLPD~\cite{rlpd} alleviates Q-function overestimation by increasing the number of critics (i.e., a larger Q-ensemble) and introducing LayerNorm in the critic networks. However, these approaches have not been extensively validated in few-shot settings or in vision-based real-world scenarios.

\subsection{Large Model-Enhanced Reinforcement Learning}
The advent of Large Language Models (LLMs) and Vision-Language Models (VLMs) has spurred interest in leveraging semantic priors for RL~\cite{wang2023voyager,yao2022react,lin2024swiftsage,jiang2022vima,shi2024yell,liu2023interactive,ouyang2024long}. Typical approaches include: (1) Reward Synthesis: Using LLMs to automatically code reward functions (e.g., Eureka~\cite{eureka2023}, T2R~\cite{t2r2024}) to bypass manual engineering; (2) Preference Learning: Employing VLMs as preference scorers (e.g., RL-VLM-F~\cite{wang2024rl}, ERL-RL~\cite{luu2025enhancing}) to provide feedback based on visual outcomes; and (3) Exploration Guidance: Leveraging LLMs to propose action priors or sub-goals (e.g., LLM-Explorer~\cite{hao2025llm}).
However, the reliability of these methods remains a critical concern. LLM-generated rewards often suffer from hallucinations or misalignment, necessitating extensive manual verification. Furthermore, VLM-based preference scoring is predominantly validated in simplified environments and often cannot distinguish between similar states in continuous-control tasks.

\subsection{Reinforcement Learning in Real Robot Scenarios}
Real-world RL tackles policy learning directly on physical systems under constraints of noise and sample efficiency. Early works focused on visual servoing~\cite{levine2016end} or large-scale grasping~\cite{kalashnikov2018qtopt}. Recent approaches streamline this process by integrating human guidance: SERL~\cite{luo2024serl} and HIL-SERL~\cite{luo2024hilserl} leverage interventions and corrections, while RL-100~\cite{rl-100} focuses on iterative self-improvement. Most recently, GR-RL~\cite{gr-rl} combines offline filtering with online RL to fine-tune VLA policies for high-precision tasks.
However, widespread deployment remains hindered by three constraints. First, heavy dependency on priors: Methods often demand dense interventions or heavy pre-trained VLA backbones (as in GR-RL), limiting usage in resource-constrained settings. Second, limited embodiment: Validations are mostly confined to tabletop manipulation, lacking cross-embodiment generalization to mobile or legged robots. Finally, complex setups: Reliance on multi-view perception or multi-stage pipelines restricts applicability in unstructured environments. In contrast, SigEnt-SAC targets a more challenging setting: learning from scratch with minimal guidance, single-view perception, and diverse robot embodiments.

\section{Notation}
We represent the environment in the standard form of a Markov Decision Process (MDP):
\[
\mdp := \langle \sspace, \aspace, \rew, \trns, \gamma, \rho \rangle .
\]
where \(\sspace\) and \(\aspace\) denote the state space and action space, respectively. Let \(\mathcal{Z} := \sspace \times \aspace\). 
The reward function
\(
\rew : \saspace \to \mathbb{R}
\)
maps each state--action pair to a scalar reward;
the transition function
\(
\trns : \saspace \to \dist(\sspace)
\)
specifies the probability distribution over next states given a state--action pair.
The initial state distribution is given by \(\rho \in \dist(\sspace)\).

A stochastic policy \( \pi:\mathcal{S} \to \Delta(\mathcal{A}) \) assigns a probability distribution over actions to each state, while a deterministic policy \( \mu:\mathcal{S} \to \mathcal{A} \) selects a single action in each state.

Let \(Q(s,a)\) denote the true Q-function of the MDP.
We define \( \hat{Q}(s,a) \) as the Q-function estimate learned via TD updates augmented with an additional regularization term;
\(\hat{Q}^{\text{soft}}(s,a)\) denotes the Q-function estimate learned under the soft Bellman operator~\cite{sac} together with an additional bias/regularization term.

Furthermore, let \(Q^*\) denote the optimal Q-function of the local MDP \(\mathcal{D}\).
We use \(\hat{Q}^*\) to denote the converged (biased) Q-function induced by the regularized TD updates.

\begin{figure}[t]
  \begin{center}
    \centerline{\includegraphics[width=\linewidth]{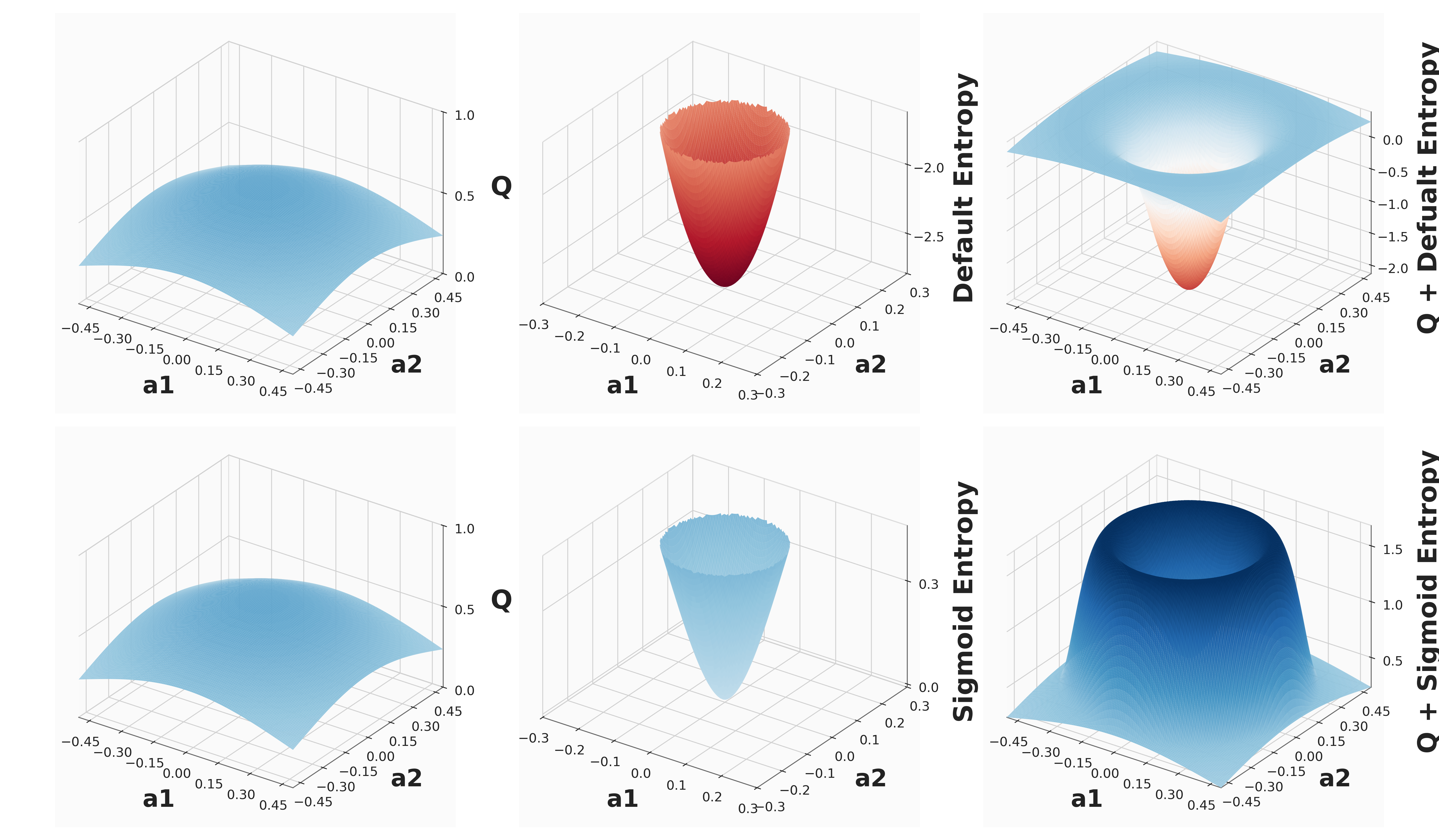}}
    \caption{Illustration of negative-entropy effects in soft Q-learning. We compute $\hat{Q}(s,a)$ assuming a Gaussian policy with $\sigma_\pi=0.1$, and assume sampled actions lie within $1.5\sigma_\pi$ of the mean. \textbf{Top row:} the default entropy term (negative entropy; row~1, col~2) lowers $\hat{Q}(s,a)$ for $a=\pi(s)$ and its neighborhood, making max-$Q$ policy improvement more likely to move toward out-of-distribution (OOD) actions. \textbf{Bottom row:} a sigmoid-bounded entropy maps the entropy contribution to a strictly positive, bounded range, yielding a clearer high-$Q$ region and a more well-defined action set for maximization.}
    \label{fig:neg_entropy_illustration}
  \end{center}
\end{figure}

\section{Sigmoid Bounded Soft Actor Critic}

The SigEnt-SAC framework is built upon two core mechanisms:

(1) \textbf{Sigmoid-Bounded Entropy}  
During training, we map the default (unbounded) entropy signal~\cite{sac} of tanh-squashed Gaussian policies to a \emph{bounded} and strictly-positive score by applying a sigmoid to the per-dimension surprisal. This bounded formulation eliminates policy over-optimization driven by negative entropy, improves the stability of Q-function updates, and mitigates the performance oscillation commonly observed in conservative Q-learning based methods.

(2) \textbf{Gated Behavior Cloning}  
During each policy update step, we introduce an expert-based gated behavior cloning loss that is added on top of the policy loss to provide additional gradients on the policy mean action. This shaping signal enables the policy to remain exploring within the neighborhood of the expert trajectory even when the Q-function is not yet converged or is oscillating, thereby reducing ineffective exploration and consequently minimizing hardware wear caused by unnecessary interactions.

Collectively, these two mechanisms mitigate the oscillation issues of conservative Q-learning based methods and improve sample efficiency, enhancing training stability for real-world deployment without large-scale pretraining.
\begin{figure}[h]
  \begin{center}
    \centerline{\includegraphics[width=\linewidth]{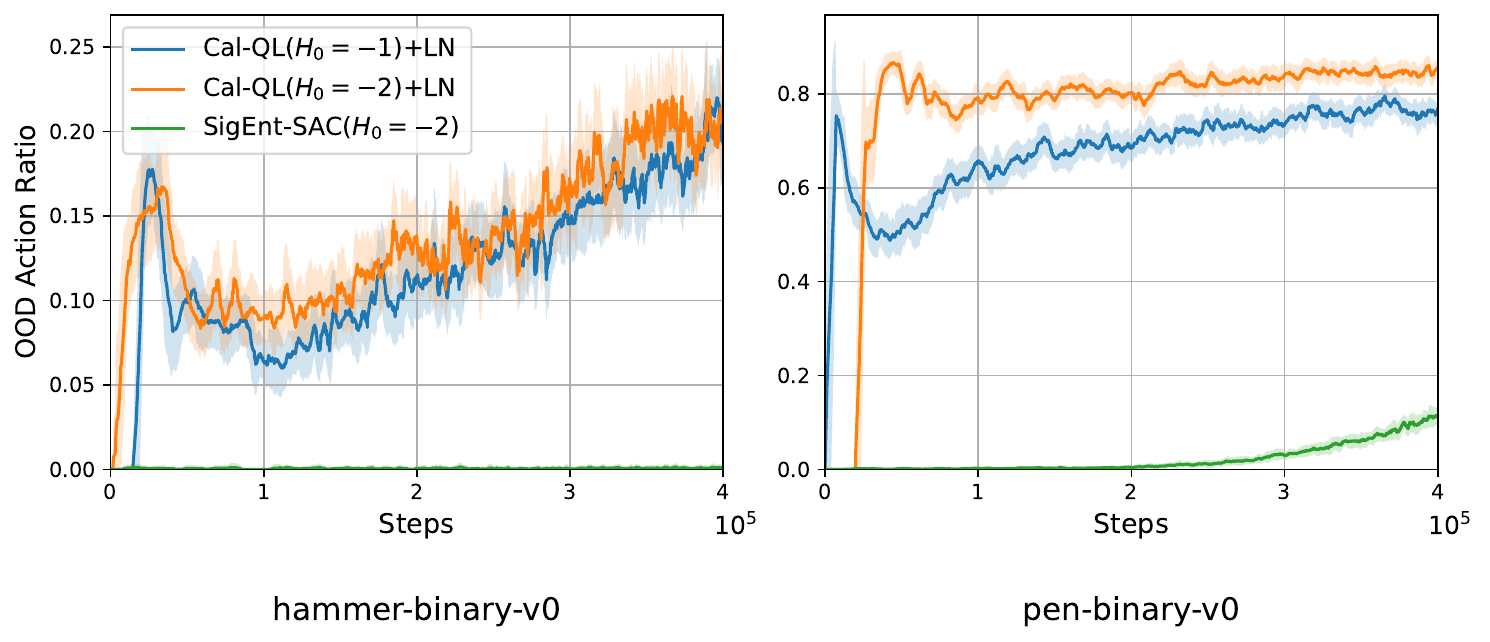}}
    \caption{Policy optimization with a negative entropy term is more likely to produce OOD actions; correspondingly, a smaller target policy entropy leads to a higher OOD action ratio. The OOD criterion is defined in Eq.~\ref{eq:p_gate} with the threshold set to 0.3. We additionally apply LayerNorm to Cal-QL to mitigate spurious OOD actions caused by network oscillations.}
    \label{fig:ood_action_presentation}
  \end{center}
\end{figure}

\subsection{Sigmoid-Bounded Entropy Mechanism}

We consider a tanh-squashed Gaussian policy $a=\tanh(x)$, where $x$ is sampled via reparameterization from the policy network outputs: $x=\mu_\theta(s)+\sigma_\theta(s)\odot\epsilon$, with $\epsilon\sim\mathcal{N}(0,I)$. Let $\log \pi_{\theta,i}(a_i|s)$ denote the \emph{per-dimension} log-density of the squashed policy (so that $\log \pi_\theta(a|s)=\sum_{i=1}^{d}\log \pi_{\theta,i}(a_i|s)$, where $d$ is the action dimension). We define the per-dimension negative log-density (surprisal) as $s_i=-\log \pi_{\theta,i}(a_i|s)$. The bounded entropy contribution is computed via a temperature-controlled sigmoid:

\begin{equation}
\label{eq:sigmoid_entropy}
h_i(s_i)=h_{\max}\cdot \sigma\!\left(\frac{s_i-m}{t}\right),
\qquad
\mathcal{H}_{\text{sig}}(s,a)=\sum_{i=1}^{d} h_i(s_i),
\end{equation}
where $\sigma(\cdot)$ is the sigmoid function, $h_{\max}>0$ controls the maximum per-dimension entropy score, $m$ is a center offset, and $t>0$ is a temperature parameter. By construction, $\mathcal{H}_{\text{sig}}(s,a)\in(0,d\cdot h_{\max})$.

Replacing the default entropy term with $\mathcal{H}_{\text{sig}}$, the Q-function update rule is formulated as:
\begin{equation}
\label{eq:q-update-sig}
\scalebox{0.95}{$
\begin{aligned}
\hat{Q}^{k+1}(s,a)
&= (1-\alpha)\,\hat{Q}^{k}(s,a)
\\
&+ \alpha \Big[
r(s,a)
+ \gamma \mathbb{E}_{a' \sim \pi}\big[\hat{Q}^{k}(s',a') + \mathcal{H}_{\text{sig}}(s',a')\big]
\Big].
\end{aligned}
$}
\end{equation}

As illustrated in Figure~\ref{fig:neg_entropy_illustration}, under the default formulation $\hat{Q}(s,a) - \alpha \log \pi(a|s)$, the negative entropy term can dominate and shift the maximum-Q region toward the action-space boundary, causing out-of-distribution exploration(see Figure~\ref{fig:ood_action_presentation}). In contrast, $\hat{Q}(s,a)+\mathcal{H}_{\text{sig}}(s,a)$ produces a ``bowl-shaped'' regularized landscape (with a central dip and low values near the boundary), which biases policy improvement toward actions within the neighborhood of the state-visit distribution $p(s)$, thereby reducing conservative-Q-learning-induced oscillations and avoiding unnecessary, hardware-costly interactions.

\subsection{Gated Behavior Cloning}

During policy updates, we integrate the learning signals derived from agent-collected transitions with the guidance provided by expert demonstrations. Distinct from conventional paradigms that combine policy optimization with Behavior Cloning (BC), we employ a dynamic gating mechanism based on the deviation between the policy mean action and the expert action. 

Let $a_{\mathrm{mean}}(s)=\tanh(\mu_\theta(s))$ denote the deterministic mean action of the policy, and define the gating mask:
\begin{equation}
\label{eq:p_gate}
p_{\mathrm{gate}}(s, a_{\mathrm{exp}}) = \mathbb{I}\!\left[ \|a_{\mathrm{mean}}(s) - a_{\mathrm{exp}}\|_2 > \varepsilon \right],
\end{equation}
where $\varepsilon > 0$ is a threshold. We then optimize a unified maximum-entropy policy objective with sigmoid-bounded entropy and a gated expert shaping term:
\begin{equation}
\label{eq:policy_obj_sig_gbc}
\scalebox{0.9}{$
\begin{aligned}
\mathcal{J}(\pi_\theta)
&=
\mathbb{E}_{s \sim \mathcal{D},\, a \sim \pi_\theta(\cdot|s)}
\Big[
Q^{k}(s,a) + \alpha\,\mathcal{H}_{\text{sig}}(s,a)
\Big]
\\
&\quad-\;
\lambda\,
\mathbb{E}_{(s,a_{\mathrm{exp}})\sim \mathcal{D}_{\mathrm{exp}}}
\Big[
p_{\mathrm{gate}}(s,a_{\mathrm{exp}})\cdot \|a_{\mathrm{mean}}(s)-a_{\mathrm{exp}}\|_2^2
\Big].
\end{aligned}
$}
\end{equation}

where $\alpha$ is the entropy weight and $\lambda>0$ controls the strength of the expert-based shaping signal.

In Equation~\ref{eq:policy_obj_sig_gbc}, the third term provides additional gradients on the policy mean action only when the deviation from the expert exceeds the threshold. This gated shaping reduces drift caused by critic oscillations and prevents unnecessary out-of-distribution exploration, while still allowing the policy to adapt beyond the demonstration.

\subsection{Algorithm Implementation}

SigEnt-SAC augments the standard off-policy actor-critic framework with two key mechanisms: \textbf{Sigmoid-Bounded Entropy} and \textbf{Gated Behavior Cloning}. The complete training procedure is outlined in Algorithm~\ref{algo:sigentsac}.

\begin{algorithm}[h]
\caption{SigEnt-SAC Training Framework}
\label{algo:sigentsac}
\begin{algorithmic}[1]
\STATE \textbf{Input:} Replay buffer $\mathcal{D}_{\mathrm{buf}}$, expert buffer $\mathcal{D}_{\mathrm{exp}}$, total training steps $N$, batch size $B$
\STATE \textbf{Initialize:} Policy parameters $\phi$, critic parameters $\theta_1, \theta_2$, and target parameters $\theta_1', \theta_2'$

\vspace{3pt}
\STATE \textcolor{gray}{\textit{// Phase 1: Expert Demonstration Collection}}
\STATE Collect expert transitions and store in $\mathcal{D}_{\mathrm{exp}}$

\vspace{3pt}
\STATE \textcolor{gray}{\textit{// Phase 2: Online Interaction \& Training}}
\FOR{$k=1$ \textbf{to} $N$}
    \STATE \textcolor{gray}{\textit{// Environment Interaction}}
    \STATE Collect transition $(s,a,r,s')$ and store in $\mathcal{D}_{\mathrm{buf}}$

    \IF{$|\mathcal{D}_{\mathrm{buf}}| \ge B$}
        \STATE Sample batch $\mathcal{B}_{\mathrm{buf}} \sim \mathcal{D}_{\mathrm{buf}}$ and expert batch $\mathcal{B}_{\mathrm{exp}} \sim \mathcal{D}_{\mathrm{exp}}$
        
        \vspace{2pt}
        \STATE \textcolor{gray}{\textit{// Critic Update (Sigmoid-Bounded Entropy)}}
        \STATE Compute $\mathcal{H}_{\text{sig}}(s',a')$ for $a' \sim \pi_\phi(\cdot|s')$ using Eq.~\ref{eq:sigmoid_entropy}
        \STATE Update critics $\theta_1, \theta_2$ using Eq.~\ref{eq:q-update-sig}
        
        \vspace{2pt}
        \STATE \textcolor{gray}{\textit{// Policy Update (Gated Behavior Cloning)}}
        \STATE Update policy $\phi$ by maximizing Eq.~\ref{eq:policy_obj_sig_gbc}
        
        \vspace{2pt}
        \STATE \textbf{Target Update:} $\theta_i' \leftarrow \tau\theta_i + (1-\tau)\theta_i'$ for $i=1,2$
    \ENDIF
\ENDFOR
\end{algorithmic}
\end{algorithm}

\paragraph{Critic Loss Function.}

The entropy-augmented Bellman target is defined as:
\begin{equation}
\label{eq:bellman_target_sig}
y = r + (1-d)\gamma \left( \min_{i=1,2} Q'_i(s', a') + \alpha\, \mathcal{H}_{\text{sig}}(s',a') \right),
\end{equation}
where $a' \sim \pi_\phi(\cdot|s')$ and $Q'$ denotes the target networks. The base TD loss is
\begin{equation}
\label{eq:td_loss}
\mathcal{L}_{\mathrm{TD}}^{(i)} = \mathbb{E}_{\mathcal{D}_{\mathrm{buf}}} \Big[\big(Q_i(s,a) - y\big)^2 \Big], \qquad i\in\{1,2\}.
\end{equation}

For each state $s$ in the replay batch, we form an OOD action set by sampling multiple actions from the current policy at both $s$ and $s'$ and concatenating them:
\[
\mathcal{A}_{\mathrm{ood}}(s)=\{a^{(j)} \sim \pi_\phi(\cdot|s)\}_{j=1}^{n}\ \cup\ \{a'^{(j)} \sim \pi_\phi(\cdot|s')\}_{j=1}^{n}.
\]
We then add a simplified CQL-style regularizer~\cite{cql}. For each critic $Q_i$:
\begin{equation}
\label{eq:cql_cap}
\scalebox{0.95}{$
\begin{aligned}
\mathcal{L}_{\mathrm{CQL}}^{(i)}
&=
\mathbb{E}_{(s,a)\sim \mathcal{D}_{\mathrm{buf}}}
\Bigg[
\beta \log \Bigg(
\sum_{\tilde{a}\in \{a\}\cup \mathcal{A}_{\mathrm{ood}}(s)}
\exp\!\Big(\tfrac{1}{\beta} Q_i(s,\tilde{a})\Big)
\Bigg)
\\
&\qquad\qquad\qquad
-\, Q_i(s,a)
\Bigg],
\end{aligned}
$}
\end{equation}
where $\beta>0$ is the log-sum-exp temperature (set to $1$ in our implementation). This term discourages spurious high Q-values on sampled OOD actions and helps reduce training oscillations. Note that $Q_i$ is still lower-bounded by its Monte-Carlo discounted return that is similar in spirit to Cal-QL~\cite{calql}.

The final critic objective for each critic $Q_i$ is:
\begin{equation}
\label{eq:critic_final_loss}
\mathcal{L}_{Q_i}
=
\mathcal{L}_{\mathrm{TD}}^{(i)}
+
\lambda_{\mathrm{ood}}\,\mathcal{L}_{\mathrm{CQL}}^{(i)},
\qquad i\in\{1,2\},
\end{equation}
where $\lambda_{\mathrm{ood}}$ controls the strength of the conservative regularization.

\paragraph{Policy Loss with Gated Behavior Cloning.}

We optimize the policy by minimizing the negative of the unified objective defined in Eq.~\ref{eq:policy_obj_sig_gbc}:
\begin{equation}
\label{eq:policy_loss_negJ}
\mathcal{L}_{\pi_\theta} = -\,\mathcal{J}(\pi_\theta).
\end{equation}

In implementation, the gated behavior cloning term is evaluated on the deterministic mean action $a_{\mathrm{mean}}(s)=\tanh(\mu_\theta(s))$, and is activated only when the per-sample deviation exceeds a threshold, i.e., $\frac{1}{d}\|a_{\mathrm{mean}}(s)-a_{\mathrm{exp}}\|_2^2 > \varepsilon_{\mathrm{bc}}^2$.

\section{Experiments}
\label{sec:experiments}

In this section, we empirically evaluate the SigEnt-SAC framework to answer the following research questions:

\begin{itemize}
    \item \textbf{Q1 (Sample Efficiency):} Can SigEnt-SAC achieve rapid convergence and high performance with extremely limited data (e.g., a single demonstration)?
    \item \textbf{Q2 (Robustness):} Is the method resilient to a suboptimal or noisy expert demonstration with varying quality?
    \item \textbf{Q3 (Real-World Adaptability \& Generalization):} Can the framework: (i) scale across diverse robotic morphologies; (ii) handle raw, unstable embodied visual observations; (iii) generalize to dynamic environmental variations; and \textbf{(iv) significantly outperform the suboptimal expert demonstration?}
    \item \textbf{Q4 (Hyperparameter Sensitivity):} How sensitive is SigEnt-SAC to the hyperparameters of the Gated Behavior Cloning (GBC) term?
    
\end{itemize}

\begin{figure*}[t]
  \begin{center}
    \centerline{\includegraphics[width=\textwidth]{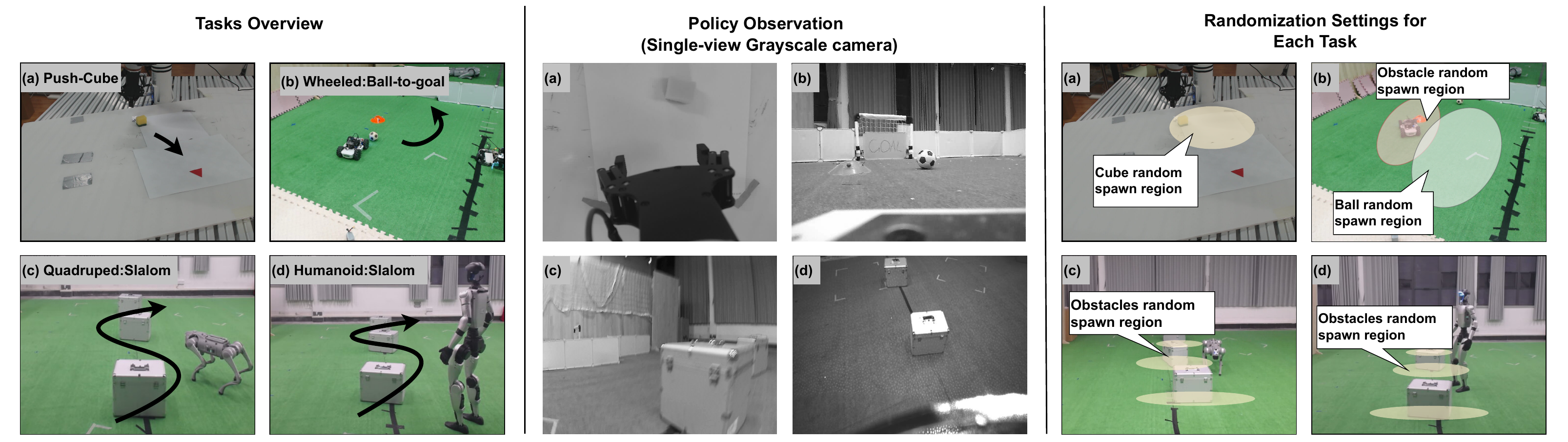}}
\caption{Real-world task suite specification. The figure shows: (Left) an overview of the tasks; (Middle) the unified policy input, consisting of a single-view grayscale local observation; (Right) per-task randomization settings: in Push Cube, the cube is initialized at random positions; in Wheel, both the obstacle and the soccer ball are randomly placed, and the obstacle is movable/collidable; in Quadruped and Humanoid, obstacles are randomly placed within a predefined region and are also movable/collidable.}

    \label{fig:real_world_generalization}
  \end{center}
\end{figure*}

\begin{figure*}[t]
  \centering
  \includegraphics[width=\textwidth]{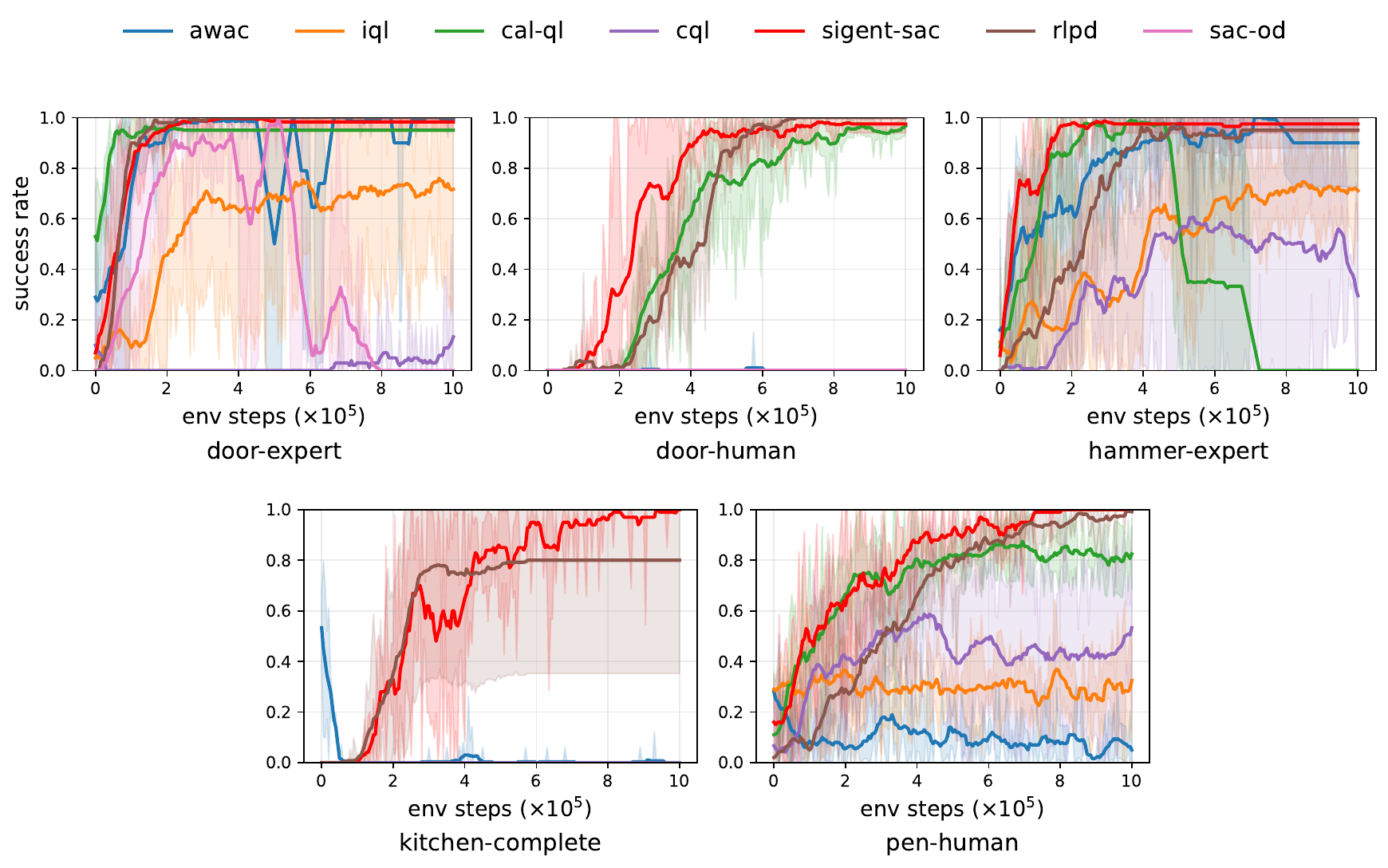}
\caption{Online learning in the one-shot setting, where agents are provided with only one demonstration and must learn to succeed within 1M steps. Compared to other baselines, \textsc{SigEnt-SAC} achieves a higher success rate and converges faster, allows the agent to explore with low target entropy, and exhibits no performance drop after convergence; in contrast, all other baselines either suffer performance drop after convergence or fail to learn throughout the entire training phase.}

  \label{fig:sim_benchmarks}
\end{figure*}

\subsection{Experimental Setup}

\paragraph{Baselines.} To evaluate the advantages of SigEnt-SAC in the offline-to-online setting, we compare it against several state-of-the-art algorithms:
(1) Cal-QL, CQL \cite{calql,cql}: conservative-based methods that penalize Q-values for out-of-distribution (OOD) actions to improve robustness in offline learning and stabilize offline-to-online fine-tuning;
(2) RLPD \cite{rlpd}, an efficient online fine-tuning method using a Q-ensemble and layernorm;
(3) AWAC \cite{awac}, a Q-value-weighted behavior cloning method that emphasizes actions with higher estimated returns;
(4) IQL \cite{iql}, an implicit Q-learning method that avoids explicit behavior constraints by learning an expectile value function and deriving the policy via advantage-weighted regression;
(5) SAC (w/ Prior), standard SAC initialized with expert data in the replay buffer.

\paragraph{Hardware Setup \& Observation Design.}
We deploy SigEnt-SAC on a diverse set of physical robotic platforms to verify cross-morphology adaptability (refer to Figure~\ref{fig:real_world_generalization}). The hardware platforms include: an AgileX LIMO Pro and a TurtleBot3 Waffle (for automated ball pickup) for vision-based ball-to-goal scoring; a RealMan RM65-B manipulator for vision-based manipulation; and a Unitree Go2 quadruped as well as a Unitree G1 humanoid for vision-based locomotion.

All platforms operate under a \textbf{purely visual, local observation} setting. The agents receive \textbf{single-view grayscale} image streams from onboard cameras without external global state estimation.

\paragraph{Task Settings.}
We evaluate our method across two domains of increasing difficulty, both utilizing sparse rewards.
1. D4RL Continuous Control: Complex manipulation tasks (Kitchen, Adroit) with only a single success trajectory provided.
2. Real-World Robotics: We design four tasks targeting the challenges in Q3: (1) Push-Cube (Manipulator), pushing a randomly placed cube into a designated goal region; (2) Ball-to-goal (Wheeled): scoring by pushing the ball into a goal, where both the obstacle and the ball are randomly positioned within a specific area and the obstacle can be pushed away; (3) Slalom (Quadruped), weaving around box-shaped obstacles, where the boxes are randomly positioned within a specific area and can be pushed away; and (4) Slalom (Humanoid), weaving around box-shaped obstacles with whole-body balance.

\begin{figure*}[th]
  \centering

  \begin{subfigure}[t]{0.49\textwidth}
    \centering
    \includegraphics[width=\textwidth]{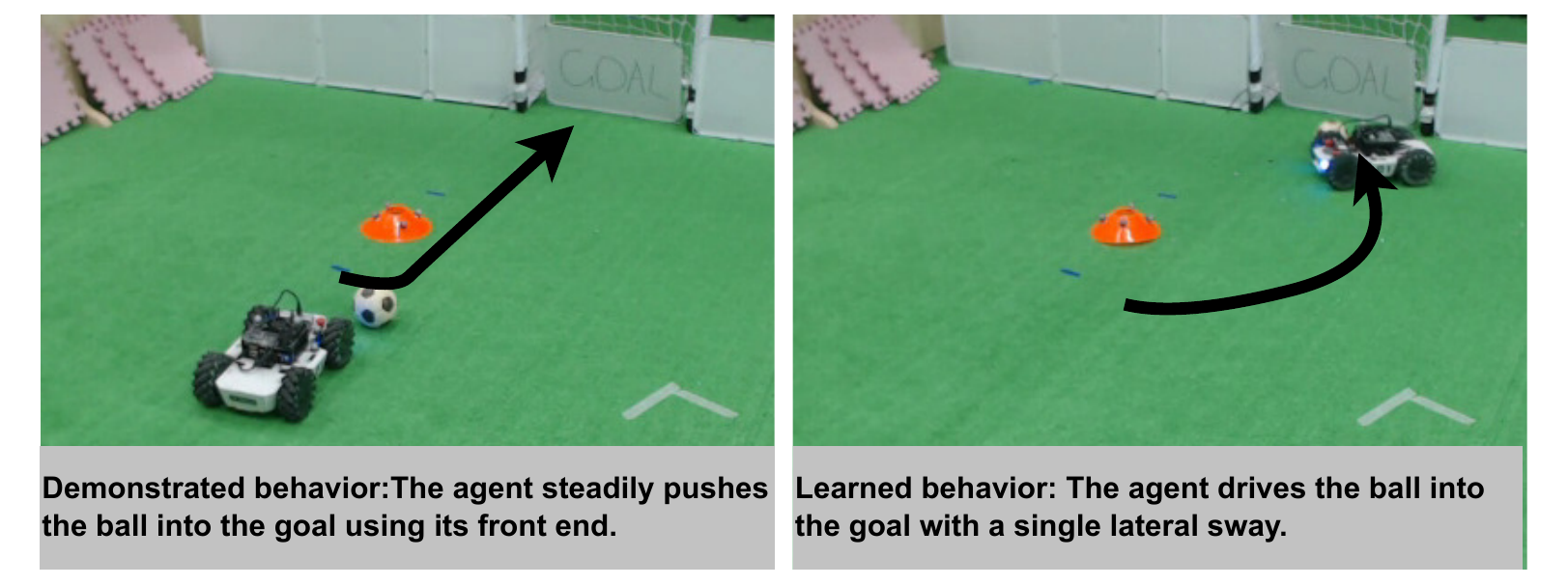}
    \caption{\textit{Ball-to-Goal} qualitative comparison.}
    \label{fig:expert_comparison}
  \end{subfigure}
  \hfill
  \begin{subfigure}[t]{0.49\textwidth}
    \centering
    \includegraphics[width=\textwidth]{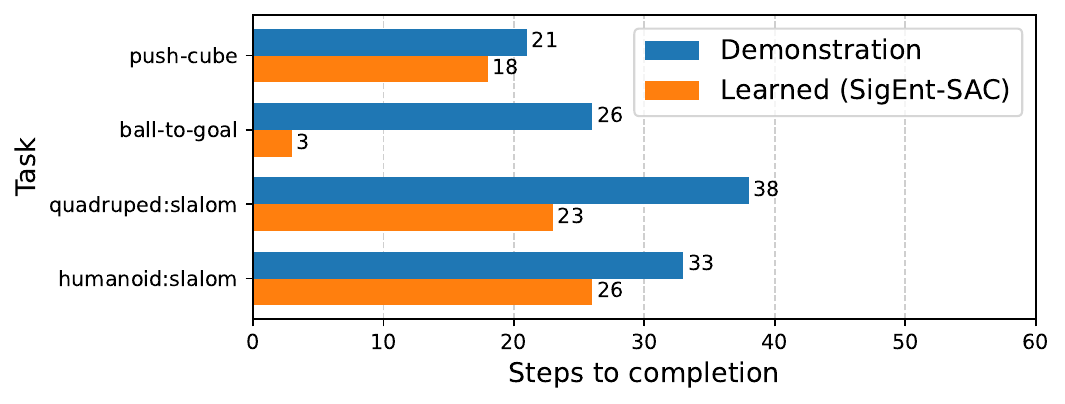}
    \caption{Steps to completion across four tasks.}
    \label{fig:steps_comparison}
  \end{subfigure}

  \caption{\textbf{Demonstration and learned policy in the real world.}
  (a) On \textit{Ball-to-Goal}, the demonstration steadily pushes the ball with the uneven front end, while SigEnt-SAC uses its flatter side and a single lateral sway to drive the ball into the goal more quickly and accurately.
  (b) Steps required to complete each task for the demonstration and the learned policy. The learned policy reduces steps by an average of 40.9\% across four tasks; on \textit{ball-to-goal}, it achieves an \textbf{88.46\%} reduction (26 $\rightarrow$ 3 steps).}
  \label{fig:realworld_demo_and_steps}
\end{figure*}

\subsection{Computational Efficiency \& Deployment Overhead}

\begin{table}[h]
\begin{center}
\begin{small}
\begin{sc}
\resizebox{\columnwidth}{!}{%
    \begin{tabular}{lccc}
    \toprule
    Method & \#Q-functions & Params (M) & Training Time (ms/update) \\
    \midrule
    AWAC & 2 & 0.215 & 0.61 \\
    IQL & 2 & 0.083 & 1.31 \\
    Cal-QL & 2 & 0.090 & 1.51 \\
    SigEnt-SAC & 2 & 0.091 & 3.13 \\
    RLPD & 10 & 0.090 & 0.78 \\
    \bottomrule
    \end{tabular}%
}
\end{sc}
\end{small}
\caption{Comparison of training and deployment overhead between SigEnt-SAC and representative baselines.}
\label{tab:efficiency}
\end{center}
\end{table}

A critical barrier to real-world RL is the policy inference cost during deployment.
As shown in Table~\ref{tab:efficiency}, we compare SigEnt-SAC with representative baselines in terms of the number of Q-functions, policy params, and training time per update.
SigEnt-SAC does not increase policy inference time but increases training cost because it introduces the gated policy cloning term to maintain policy stability.

\subsection{Simulation Evaluation (Q1 \& Q2)}

\begin{figure}[t]
  \begin{center}
    \centerline{\includegraphics[width=\linewidth]{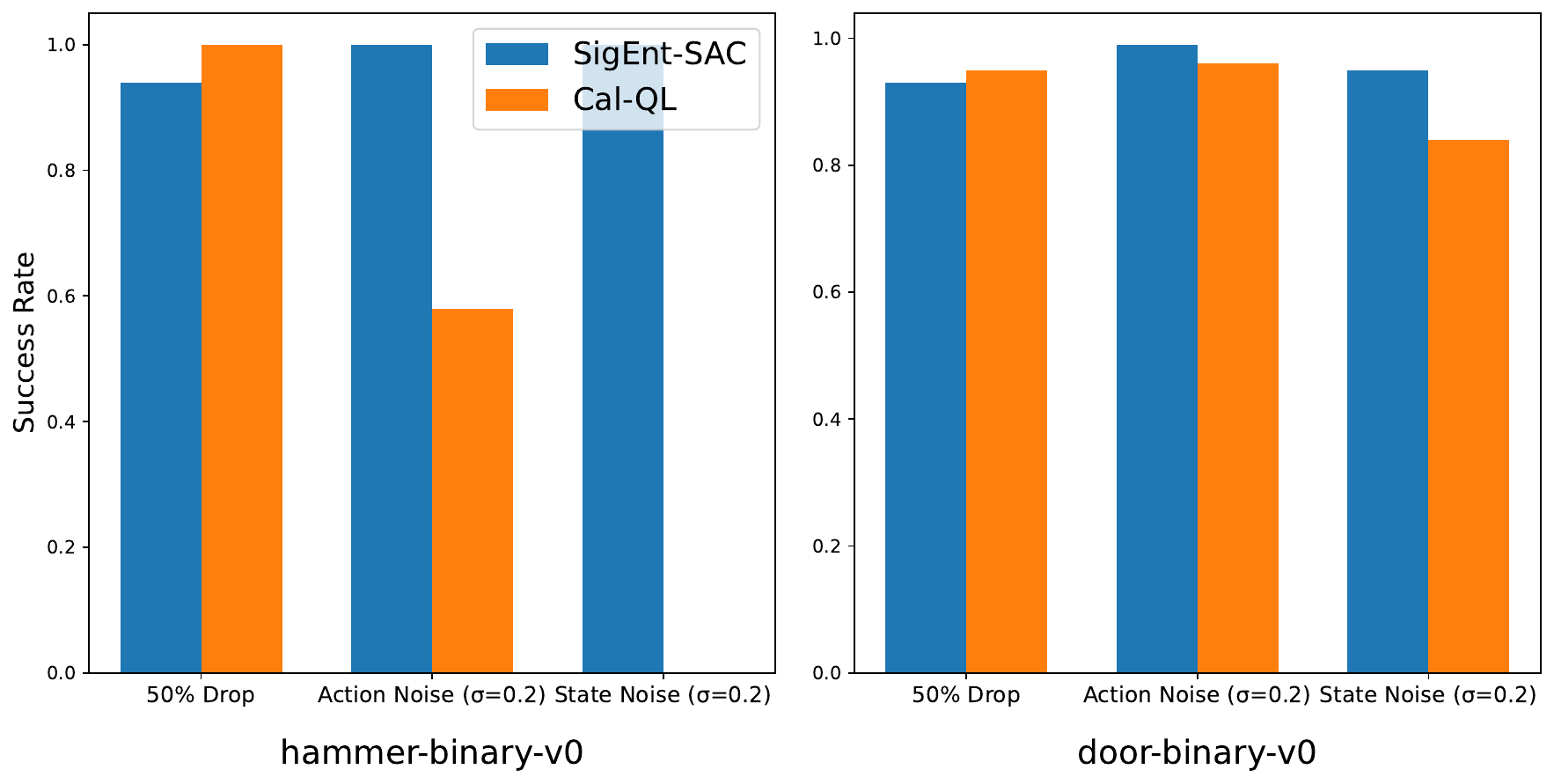}}
    \caption{Robustness to demonstration quality degradations. We evaluate learning performance under three one-shot demonstration variants: 50\% drop, action noise ($\sigma=0.2$), and state noise ($\sigma=0.2$). SigEnt-SAC is consistently robust to noisy demonstrations, while Cal-QL suffers substantial performance degradation, most notably on the hammer task.}
    \label{fig:data_quality}
  \end{center}
\end{figure}

\paragraph{SigEnt-SAC Is More Robust to Noisy Demonstrations.}
We study how the quality of one-shot expert demonstrations affects learning performance, aiming to reduce reliance on carefully curated expert data.
We construct three degraded demonstration variants: (1) a successful trajectory with 50\% of transitions dropped; (2) the same trajectory corrupted by Gaussian noise on actions with $\sigma = 0.2$; and (3) corrupted by Gaussian noise on states with $\sigma = 0.2$.
As shown in Figure~\ref{fig:data_quality}, SigEnt-SAC exhibits stronger robustness than Cal-QL under noisy demonstration settings. In contrast, Cal-QL suffers a pronounced performance drop when the demonstration is corrupted by noise, which we attribute primarily to its large target-entropy setting ($H_0 = 0$), which induces higher policy stochasticity, making learning more sensitive to imperfect (noisy) guidance.

\paragraph{SigEnt-SAC Reaches 100\% Success Faster in the One-Shot Setting.}

To evaluate our proposed method in the one-shot setting, we conduct experiments on high-dimensional D4RL tasks and compare against state-of-the-art baselines. Figure~\ref{fig:sim_benchmarks} shows the learning curves. SigEnt-SAC reaches a 100\% success-rate policy faster than all baselines and yields more stable Q-value estimates than conservative methods.

\subsection{Real-World Robotic Evaluation (Q3)}

We evaluate SigEnt-SAC on four physical platforms across four real-world tasks. In all scenarios, only one successful trajectory is used as the offline dataset.

\paragraph{SigEnt-SAC Is Robust to Embodied Visual Observations.}
SigEnt-SAC consistently converges across all four tasks under a purely visual, single-view observation setting, demonstrating reliable offline-to-online learning with minimal supervision (one demonstration).
As shown in Table~\ref{tab:real_world_success}, we compare SigEnt-SAC against two representative baselines: (i) behavior cloning (BC) implemented with a Vision Transformer policy~\cite{vit} trained on 30 demonstrations, and (ii) a zero-shot VLM agent using ChatGPT-5.2. 
We find that BC struggles under limited data and lacks a recovery mechanism, leading to low success rates across tasks. 
The zero-shot VLM agent frequently fails to interpret the control command and cannot produce sufficiently accurate actions, resulting in near-zero success. 
In contrast, SigEnt-SAC achieves 100\% success on all tasks with only a single demonstration.

\paragraph{SigEnt-SAC Surpasses the Demonstration.}
Through online data collection and policy improvement, SigEnt-SAC learns faster strategies than the provided trajectory, achieving an average \textbf{40.9\%} reduction in task completion time (Figure~\ref{fig:realworld_demo_and_steps}).

\begin{table}[t]
\begin{center}
\begin{small}
\begin{sc}
\begin{tabular}{lccc}
\toprule
Task & BC & VLM & \textbf{SIGENT-SAC} \\ 
\midrule 
Push-Cube (Manipulator) & 20\% & 10\% & \textbf{100\%} \\ 
Ball-Driving (Wheeled) & 10\% & 10\% & \textbf{100\%} \\ 
Slalom (Go2) & 0\% & 0\% & \textbf{100\%} \\ 
Slalom (G1) & 0\% & 0\% & \textbf{100\%} \\ 
\bottomrule 
\end{tabular} 
\end{sc} 
\end{small} 
\end{center} 
\caption{Real-world success rates (\%) over 10 evaluation trials per task. We compare BC, VLM, and SIGENT-SAC.}
\label{tab:real_world_success}
\end{table} 
 
\begin{figure}[th]
  \begin{center}
    \centerline{\includegraphics[width=\linewidth]{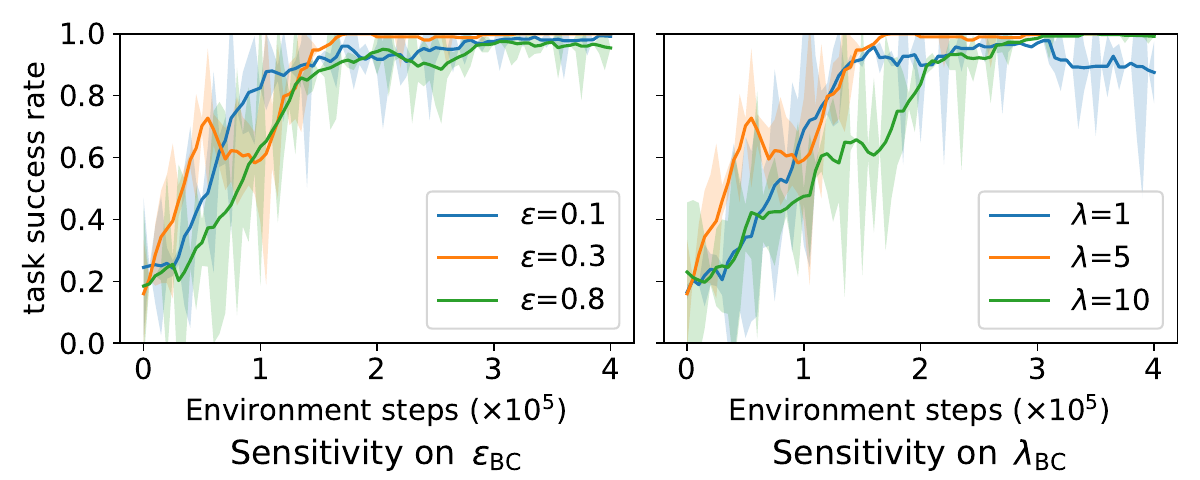}}
    \caption{Hyperparameter sensitivity analysis. We analyze the sensitivity of the boosting coefficient $\lambda$ and the gating threshold $\epsilon$. The method exhibits a wide stable region, indicating ease of tuning.}
    \label{fig:hyperparam_sensitivity}
  \end{center}
\end{figure}

\subsection{Sensitivity Analysis (Q4)}

To assess the robustness of \textbf{Gated Behavior Cloning}, we analyze the sensitivity of its key hyperparameters.

\paragraph{Sensitivity to the GBC Weight $\lambda$.}
We vary the coefficient $\lambda$ in Eq.~\ref{eq:policy_obj_sig_gbc} to control the strength of the expert-based shaping term (Figure~\ref{fig:hyperparam_sensitivity}). 
When $\lambda$ is too small, the expert signal becomes weak and the policy relies primarily on online RL updates; in the one-shot setting this leads to more oscillatory behavior as the agent lacks a stable expert-derived correction. 
As $\lambda$ increases to a moderate range, the gated shaping provides helpful corrective gradients when the policy deviates from the expert, improving sample efficiency and accelerating convergence. 
However, overly large $\lambda$ can over-regularize the policy toward the demonstration, hindering exploration and adaptation beyond the expert; as a result, convergence can become slower than with smaller (but still effective) $\lambda$ values.

\paragraph{Sensitivity to the Gating Threshold $\varepsilon$.}
We further analyze the gating threshold $\varepsilon$ in Eq.~\ref{eq:p_gate}, which determines when the expert shaping is activated (Figure~\ref{fig:hyperparam_sensitivity}). 
With a very small $\varepsilon$, the gate is frequently open, making the update resemble unconditional behavior cloning; this can slow convergence because the policy is forced to closely fit potentially suboptimal demonstration trajectories, limiting effective improvement beyond the expert. 
With a very large $\varepsilon$, the gate rarely activates, weakening the expert guidance and reducing the benefits of demonstration in the early stage; consequently, early exploration becomes more susceptible to Q-function oscillations, increasing the amount of ineffective exploration. 
Overall, SigEnt-SAC remains robust across a broad range of $\lambda$ and $\varepsilon$, reducing the burden of hyperparameter tuning when transferring to new real-world tasks.

\section{Limitations and Conclusion}
\paragraph{Limitations}
We acknowledge certain limitations in our current study. First, our control interface is restricted to high-level velocity commands, and we do not yet demonstrate full-body, joint-level control. Second, our real-world evaluation scenarios remain limited: we primarily focus on short-horizon, coarse tasks, and have not validated SigEnt-SAC on finer-grained manipulation or long-horizon control problems. In future work, We plan to further validate the effectiveness of SigEnt-SAC in more diverse settings and longer-horizon tasks.

\paragraph{Conclusion}
In this paper, we presented SigEnt-SAC, a more stable conservative-based Q-learning method that enables policy learning and convergence in real-world settings under highly constrained supervision, using only a single demonstration and a single onboard camera view. Compared to other offline-to-online baselines, SigEnt-SAC finds a 100\% success-rate policy faster, and generalizes across diverse embodiments: it is effective for both low-speed manipulation tasks and high-frequency locomotion tasks.

\bibliographystyle{named}
\bibliography{references}

\end{document}